# Application of Statistical Features in Handwritten Devnagari Character Recognition

S. Arora[1], D. Bhattacharjee[2], M. Nasipuri[2], D.K. Basu[2], M.Kundu[2]
[1]Department of CSE&IT, Meghnad Saha Institute of Technology, Kolkata-107, India
Email: sandhyabhagat@yahoo.com
[2]Department of Computer Science and Engg, Jadavpur University,kolkata,India
Email: debotoshb@hotmail.com , mitanasipuri@yahoo.com,dipakkbasu@gmail.com,mkkundu@cse.jdvu.ac.in

*Abstract*—**In this paper a scheme for offline Handwritten Devnagari Character Recognition is proposed, which uses different feature extraction methodologies and recognition algorithms. The proposed system assumes no constraints in writing style or size. First the character is preprocessed and features namely : Chain code histogram and moment invariant features are extracted and fed to Multilayer Perceptrons as a preliminary recognition step. Finally the results of both MLP's are combined using weighted majority scheme. The proposed system is tested on 1500 handwritten devnagari character database collected from different people. It is observed that the proposed system achieves recognition rates 98.03% for top 5 results and 89.46% for top 1 result.**

## I. INTRODUCTION

India is a multi-lingual and multi-script country comprising of eleven different scripts. Devnagari is third most widely used script, used for several major languages such as Hindi, Sanskrit, Marathi and Nepali, and is used by more than 500 million people. But not much work has been done towards off-line handwriting recognition of Devnagari script.Although first research report on handwritten Devnagari characters was published in 1977 but not much research work is done after that. Sinha and Mahabala 1979 [7], Veena 1999 [8], Pal and chaudhuri 1997[9] reported work on handwritten and printed Devnagari. At present researchers have started working on handwritten Devnagari characters and few research reports are available Devnagari numeral recognition [4] and towards Devnagari offline handwritten character recognition[1,2,3,5,6] after year 1977.

Bhattacharya et al [4] proposed a Multi-Layer Perceptron (MLP) neural network based classification approach for the recognition of Devnagari handwritten numerals and obtained 91.28% results. N. Sharma and U. Pal [1] proposed a directional chain code features based quadratic classifier and obtained 80.36% accuracy for handwritten Devnagari characters and 98.86% accuracy for handwritten Devnagari numerals. In most of the works reported above, multiple classifier combination has not been reported for handwritten Devnagari characters. Most of them are based on single classifier or reported for handwritten Devnagari numerals. In this paper we present the results of multiple classifier combination designed on different features for offline handwritten Devnagari character recognition. Extracted features are based on chain code histogram and moment based. Chain codes histogram features are extracted from scaled contour of the image. Moment features are extracted from scaled and thinned character image. These features are then fed to the Multi layer Perceptron for recognition.

Rest of the paper is organized as follows. In section II, characteristics of Devnagari script are discussed. Feature extraction techniques are in section III. Section IV deals with the classifiers used for the recognition purpose. The experimental results are discussed in section V.

## II. CHARACTERISTICS OF DEVNAGARI SCRIPT

Devnagari script is different from Roman script in several ways. This script has two-dimensional compositions of symbols: core characters in the middle strip, optional modifiers above and/or below core characters. Two characters may be in shadow of each other. While line segments (strokes) are the predominant features for English, most of the characters in Devnagari script is formed by curves, holes, and also strokes. In Devnagari language scripts, the concept of upper-case, the lower-case characters is absent.It consists of 14 vowels and 33 consonants. Vowels occur either in isolation or in combination with consonants. Apart from vowels and consonants characters called basic characters, there are compound characters in Devnagari script, which are formed by combining two or more basic characters. Coupled to this in Devnagari script there is a practice of having twelve forms of modifiers with each for 33 consonants , giving rise to modified shapes which, depending on whether the modifier is placed to the left, right, top or bottom of the character. The net result is that there are several thousand different shapes or patterns, which makes Devnagari OCR more difficult to develop.

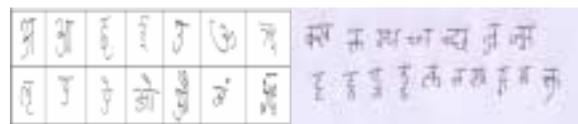

(a)                    (b)

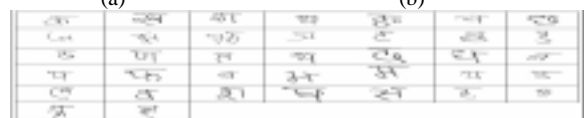

( c )

Figure 1. Sample of Handwritten Devnagari a) vowel b) compound characters c) consonants

## III. FEATURE EXTRACTION

In this section we give a brief description of the feature sets used in our proposed multiple classifier system. Chain code histogram features are extracted by chain coding the contour points of the scaled character bitmapped image. Moment based features are extracted from scaled, thinned one pixel wide skeleton of character image.

### A. Chain Code Histogram of Character Contour

Given a scaled binary image, we first find the contour points of the character image. We consider a 3 × 3 window surrounded by the object points of the image. If any of the 4-connected neighbor points is a background point then the object point (P), as shown in Fig.2 is considered as contour point.

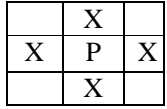

Figure 2. Contour point detection

The contour following procedure generates a contour representation called "chain coding" that is used for contour following proposed by Freeman, shown in Fig. 3a. Each pixel of the contour is assigned a different code that indicates the direction of the next pixel that belongs to the contour in some given direction. In this methodology of using a chain coding of connecting neighboring contour pixels, the points and the outline coding are captured. Contour following procedure may proceed in clockwise or in counter clockwise direction. Here, we have chosen to proceed in a clockwise direction.

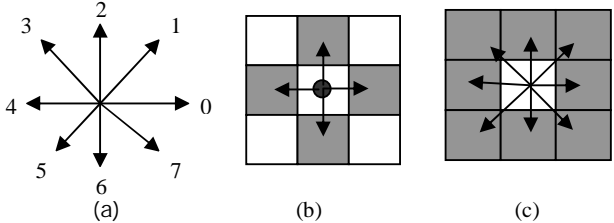

Figure 3. Chain Coding: (a) direction of connectivity, (b) 4-connectivity, (c) 8-connectivity. Generate the chain code by detecting the direction of the next-in-line pixel

The chain code for the character contour will yield a smooth, unbroken curve as it grows along the perimeter of the character and completely encompasses the character. When there is multiple connectivity in the character, then there can be multiple chain codes to represent the contour of the character. We chose to move with minimum chain code number first. We divide the contour image in 5 × 5 equal sub-images. In each of these sub-images, the frequency of the 8-way direction code is computed and a histogram of chain codes is prepared for each block. Thus we get 5 × 5 × 8 = 200 chain code features for recognition.

### B. Moment based features

Region moment representations interpret a normalized gray-level image function as a probability density of a 2D random variable. Assuming that non-zero pixel values represent regions, moments can be used for binary or gray-level transformations. Translation invariance can be achieved by using the central moments. For a digital image the central moments can be expressed as:-

$$\mu_{pq} = \sum_x \sum_y (x - \bar{x})^p (y - \bar{y})^q f(x,y)$$

where $\bar{x}$, $\bar{y}$ are the co-ordinates of the region's center of gravity (centroid). These can be obtained using the following equations:

$$\bar{x} = \frac{m_{10}}{m_{00}} \text{ and } \bar{y} = \frac{m_{01}}{m_{00}}$$

The central moments of up to order 3 can be obtained from the above equation by choosing $p, q = 0, 1, 2, 3$ such that $p + q \leq 3$. The normalized central moments denoted by $\eta_{pq}$, are denoted by $\eta_{pq} = \mu_{pq} / \mu_{00}^y$ where $y = (p + q)/2 + 1$ for $p + q = 2, 3...$. Rotation invariance can be achieved if the coordinate system is chosen such that $\mu_{11} = 0$. Seven rotation, translation, and scale invariant moment characteristics can be derived from the second and third moments.

$\phi_1 = \eta_{20} + \eta_{02}$

$\phi_2 = (\eta_{20} - \eta_{02})^2 + 4\eta_{11}^2$

$\phi_3 = (\eta_{30} - 3\eta_{12})^2 + (3\eta_{21} - \eta_{03})^2$

$\phi_4 = (\eta_{30} + \eta_{12})^2 + (\eta_{21} + \eta_{03})^2$

$\phi_5 = (\eta_{30} - 3\eta_{12})(\eta_{30} + \eta_{12})[(\eta_{30} + \eta_{12})^2 - 3(\eta_{21} + \eta_{03})^2] + (3\eta_{21} - \eta_{03})(\eta_{21} + \eta_{03})[3(\eta_{30} + \eta_{12})^2 - (\eta_{21} + \eta_{03})^2]$

$\phi_6 = (\eta_{20} - \eta_{02})[(\eta_{30} + \eta_{12})^2 - (\eta_{21} + \eta_{03})^2] 4\eta_{11}(\eta_{30} + \eta_{12})(\eta_{21} + \eta_{03})$

$\phi_7 = (3\eta_{21} - \eta_{03})(\eta_{30} + \eta_{12})[(\eta_{30} + \eta_{12})^2 - 3(\eta_{21} + \eta_{03})^2] + (3\eta_{21} - \eta_{03})(\eta_{21} + \eta_{03})[3(\eta_{30} + \eta_{12})^2 - (\eta_{21} + \eta_{03})^2]$

The values of these seven moments for a given basic symbol image represent the basic symbol and are used to create a feature vector consisting of seven values. Image is segmented into nine equal sub-images and in each sub-images moment features are calculated so total 63 features are formed.

## IV. CHARACTER RECOGNITION

We used different MLPs with 3 layers including one hidden layer for two different feature sets consisting of 200 chain code histogram features and 63 moment based features for recognition of Devnagari characters. In this work, we have considered only isolated basic non-compound Devnagari characters. Each MLP is trained with Backpropagation learning algorithm. It minimizes the sum of squared errors for the training samples by conducting a gradient descent search in the weight space. As activation function we used sigmoid function. Learning rate and momentum term are set to 0.8 and 0.7 respectively. Numbers of neurons in input layer of MLPs are 200 and 63 for chain code histogram and momentum based respectively. Number of neurons in Hidden layer is not fixed, we experimented on the values between 20-50 to get optimal result and finally it was set to 50 and 45 for chain code histogram and moment based features respectively. The output layer contained one node for each class, so the number of neurons in output layer is 20.

### A. Classifier Combination

The ultimate goal of designing pattern recognition system is to achieve the best possible classification performance. Different classifier designs potentially offered complementary information about the pattern to be classified which could be harnessed to improve the performance of the selected classifier. So instead of relying on a single decision making scheme we can combine classifiers. We have two Neural networks classifiers as discussed above, which are trained on 200 chain code and 63 moment based features respectively. The outputs are confidences associated with each class. As these outputs cannot be compared directly, we used an aggregation function for combining the results of these two classifiers. Our strategy is based on weighted majority voting scheme as described below:-

If $k^{th}$ classifier decision to assign the unknown pattern to the $i^{th}$ class is denoted by $O_{ik}$ with $1 \leq i \leq m$, m being the number of classes, then the final combined decision $d_i^{cm}$ supporting an unknown pattern assignment to the $i^{th}$ class will be represented as:

$$d_i^{com} = \sum_{k=1,2} \omega_k * O_{ik} \quad 1 \leq i \leq m$$

The final decision $d^{com}$ is therefore $d^{com} = \max d_i^{com}$, $1 \leq i \leq m$

$$\omega_k = d_k \Big/ \sum_{k=1}^{2} d_k$$

where $d_k$ is the success rate of k-th classifier. Here, classifier1 uses chain code histogram feature and has a recognition success rate $d_1$=88.19% and classifier 2 uses moment based features with success rate $d_2$=65.67%. Thus, we get $\omega 1$=0.574 and $\omega 2$=0.426.

## V. RESULTS

The experimental evaluation of the above technique was carried out for a total of 1500 samples of Devnagari basic characters (vowels as well as consonants), out of which 65% characters are used for the training and rest is used for testing purpose. The recognition accuracy obtained from chain code histogram based classifier is 88.19% and moment based classifier is 65.67%. MLP's Combined using weighted majority scheme is giving 98.03% accuracy as we considered top 5 choices results and 89.46% as top 1 choices results. We applied 3-fold cross validation testing and compared our current results with those existing pieces of work. Details comparative results are given in Table II. From experiment we noticed that mainly the error occurred because of the similar shaped characters. Some similar shape pair of Devnagari Characters which are correctly classified and misclassified most of the time in our experiment are shown in Table I

Table I. Classified and confused characters of experiment

| Characters correctly classified | Confused characters |
|---|---|
| ब स ग न न न त न ड ह | द ढ,घा, क फ,ठड,चय |

Table II. Comparison of Results

| S.N. | Method purposed by | Accuracy |
|---|---|---|
| 1. | Kumar and Singh [2] | 80% |
| 2. | N. Sharma, U. Pal, F. Kimura, and S. Pal [1] | 80.36% |
| 3. | M. Hanmandlu, O.V. R. Murthy, V.K. Madasu[3] | 90.65% |
| 4. | S. Arora, D. Bhattacharjee, M. Nasipuri, [6] | 92.80% |
| 5. | U. Pal, N. Sharma, T. Wakabayashi and F. Kimura [5] | 95.13% |
| 6. | Proposed method | 98.03% |

## VI. CONCLUSION

In this paper we present a technique of recognition of offline handwritten Devnagari characters using MLP. In future we plan to experiment on other feature extraction methods to get higher recognition accuracy.

ACKNOWLEDGMENT

Authors are thankful to the "Centre for Microprocessor Application for Training Education and Research" and "Project on Storage Retrieval and Understanding of Video for Multimedia", at Jadavpur University, Kolkata for providing the necessary facilities for carrying out this work. First author gratefully acknowledge the support of the MSIT for carrying out this research work.